\documentclass[lettersize,journal]{IEEEtran}
\usepackage{amsmath,amsfonts}
\usepackage[noend]{algorithmic}
\usepackage{algorithm}
\usepackage{array}
\usepackage[caption=false,font=normalsize,labelfont=sf,textfont=sf]{subfig}
\usepackage{textcomp}
\usepackage{stfloats}
\usepackage{url}
\usepackage{verbatim}
\usepackage{graphicx}
\usepackage{cite}
\usepackage{booktabs} 
\usepackage{multirow}
\usepackage{xcolor}
\usepackage[super]{nth}
\usepackage{tikz}

\newcommand*\circledB[1]{\tikz[baseline=(char.base)]{
            \node[shape=circle,fill,inner sep=0.2pt] (char) {\textcolor{white}{#1}};}}
\usepackage{soul}
\usepackage{enumitem}

\usepackage[switch]{lineno}

\begin{document}


\title{QuBLAST: A Framework for Quantizing Large Language Models with Block-Level Compression Approach and Activation Scaling Strategy}

\author{Pasindu Wickramasinghe$^*$, Achyuta Muthuvelan$^*$, Rachmad Vidya Wicaksana Putra$^*$,~\IEEEmembership{Member,~IEEE,}, Minghao Shao, and Muhammad Shafique,~\IEEEmembership{Senior Member,~IEEE} 
\thanks{Pasindu Wickramasinghe, Achyuta Muthuvelan, Rachmad Vidya Wicaksana Putra, and  Minghao Shao are with eBRAIN Lab, Division of Engineering, New York University (NYU) Abu Dhabi, United Arab Emirates; 
(e-mail: pmw6287@nyu.edu, am12729@nyu.edu, rachmad.putra@nyu.edu, shao.minghao@nyu.edu). \\
\indent Muhammad Shafique is the Director of eBRAIN Lab, Division of Engineering, New York University (NYU) Abu Dhabi, United Arab Emirates; 
(e-mail: muhammad.shafique@nyu.edu). \\
$^*$ Equal Contribution}
}


\maketitle

\begin{abstract}
Large Language Models (LLMs) have become the state-of-the-art algorithms for solving natural language processing tasks. 
However, they typically come at huge computational and memory costs, thus making them difficult to deploy on embedded systems. 
Toward this, state-of-the-art methods typically employ uniform post-training quantization (PTQ) across attention blocks of the network, hence overlooking the potential of applying different quantization levels in the same network.   
They also employ complex operations to mitigate the negative impact of activation outliers, hence incurring high computational overheads. 
Moreover, they have not considered evaluation using emerging LLMs with non-conventional attention architectures (e.g., state-space models), which pose different challenges in applying quantization.
To address these limitations, we propose \textit{QuBLAST}, a novel PTQ methodology that employs block-level compression approach with activation scaling strategy for LLMs.
Block-level compression approach enables mixed-precision quantization across blocks of the network, while activation scaling strategy efficiently mitigates the negative impact of activation outliers.  
Specifically, QuBLAST first analyzes the sensitivity of different attention blocks in the pre-trained model through the cross-entropy loss analysis. 
QuBLAST leverages this sensitivity analysis to determine the weight quantization level for each attention block in the model. 
Furthermore, QuBLAST employs the activation scaling map for each block to control the range of activation values and mitigate the negative impact of activation outliers, thereby enabling better quantization results.
Experimental results show that, QuBLAST reduces model sizes by 40\%-45.2\% across different model architectures (i.e., Qwen3-8B, Llama3-8B, Mistral v0.1-8B, and Falcon H1R-7B), while maintaining the performance within 5\% perplexity increase for the WikiText-2 and WikiText-103 datasets.
These highlight the effectiveness of our QuBLAST in quantizing different LLMs with different attention architectures while preserving high performance, thereby enabling their embedded deployments.   
\end{abstract}


%
\IEEEpeerreviewmaketitle

\section{Introduction}
\label{Sec_Intro}

\begin{figure}[t]
\centering
\includegraphics[width=\linewidth]{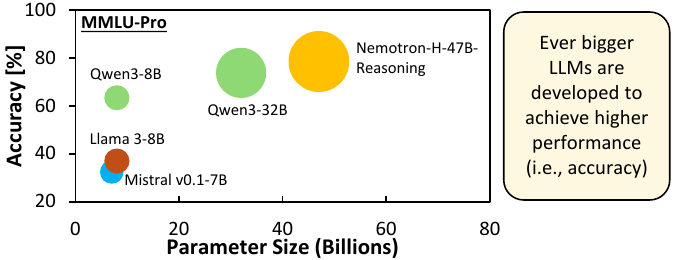}
\vspace{-0.6cm}
\caption{Current trends of the performance (i.e., accuracy) and number of parameters of different LLMs considering the MMLU-Pro dataset~\cite{wang2024mmlu}.}
\label{Fig_Trends}
\vspace{-0.3cm}
\end{figure}

Transformer-based Large Language Models (LLMs)~\cite{Ref_Vaswani_Attention_NIPS17} have demonstrated high performance in enhancing the capabilities of natural language models for providing high-quality language understanding and responses to the inputs.
Therefore, LLMs have become the state-of-the-art neural network (NN)-based algorithms for solving many natural language processing (NLP) tasks~\cite{Ref_Zhao_LLMsurvey_arXiv23, Ref_Minaee_LLMsurvey_arXiv24, Ref_Chang_LLMsurvey_TIST24, Ref_Shao_SurveyLLM_Access}.  
Due to their high performance, adoption of LLMs in diverse embedded applications is highly favored and actively being pursued for enabling LLM-powered personalized agentic systems with better privacy~\cite{Ref_ElMir_LLM4Healthcare_ICIPCW24}.
However, to achieve higher performance, ever bigger LLMs are developed, as illustrated in Fig.~\ref{Fig_Trends}. 
Consequently, LLMs have huge computational and memory demands, which make their embedded deployments difficult and non-trivial. 
Toward this, model compression is the potential solution for reducing the computational and memory requirements of LLMs.
Specifically, quantization is one of the prominent methods for compressing NN-based models~\cite{Ref_Shao_SurveyLLM_Access}\cite{Ref_Putra_FSpiNN_TCAD20, Ref_Putra_QSpiNN_IJCNN21, putra2025qsvit, putra2026qslm}, since it effectively reduces memory footprints with negligible performance degradation.
However, developing an effective quantization method is challenging, as aggressive quantization can significantly degrade the model performance.

Therefore, \textbf{the targeted research problem} in this paper is \textit{how can we perform effective quantization for any given pre-trained LLMs considering different architectural variants, while maintaining their high performance (e.g., perplexity)?} 
A solution to this problem further advances the studies for enabling tinyLLM and its efficient embedded deployments.

\subsection{State-of-the-art of LLM Quantization and Their Limitations}
\label{Sec_Intro_SOTA}
 
Many LLM quantization methods have been proposed in the literature, and the prominent ones are discussed below.
LLM-QAT~\cite{Ref_Liu_LLMQAT_ACL24} leverages quantization-aware training (QAT) approach, but at the cost of high computational, memory, and power/energy overheads for long training. 
Another approach considers post-training quantization (PTQ) to avoid excessive overheads from QAT, such as the following.
ZeroQuant~\cite{Ref_Yao_ZeroQuant_NeurIPS22} uses distillation for regaining performance loss after quantization; 
GPTQ~\cite{Ref_Frantar_GPTQ_arXiv22} quantizes weights based on Hessian analysis;
AWQ~\cite{Ref_Lin_AWQ_MLSys24} quantizes most of the weights while keeping the salient ones in higher precision; 
SmoothQuant~\cite{Ref_Xiao_SmoothQuant_ICML23} moves the difficulties of activation quantization to weights;
QLLM~\cite{Ref_Qiu_QLLM_ICLR24} reallocates the magnitude of outliers to other channels;
OmniQuant~\cite{Ref_Shao_OmniQuant_ICLR24} performs quantization error minimization; 
while SpinQuant~\cite{Ref_Liu_SpinQuant_ICLR25} employs a rotation matrix to reduce outliers and improve quantizability.
These methods have demonstrated their immense benefits for compressing LLMs, while preserving LLMs' high performance.
However, these methods still have several limitations, as the following.
\begin{itemize}[leftmargin=*]
    \item They typically employ a uniform quantization level across all attention blocks of the model, hence leading to suboptimal memory savings.
    \item They employ complex operations to mitigate the negative impact of activation outliers, thereby incurring high computational overheads.
    \item Moreover, they have not considered emerging LLMs with non-conventional attention architectures (e.g., state-space models) in their quantization techniques, which also pose different challenges in applying quantization.
\end{itemize}

\begin{figure}[t]
\centering
\includegraphics[width=\linewidth]{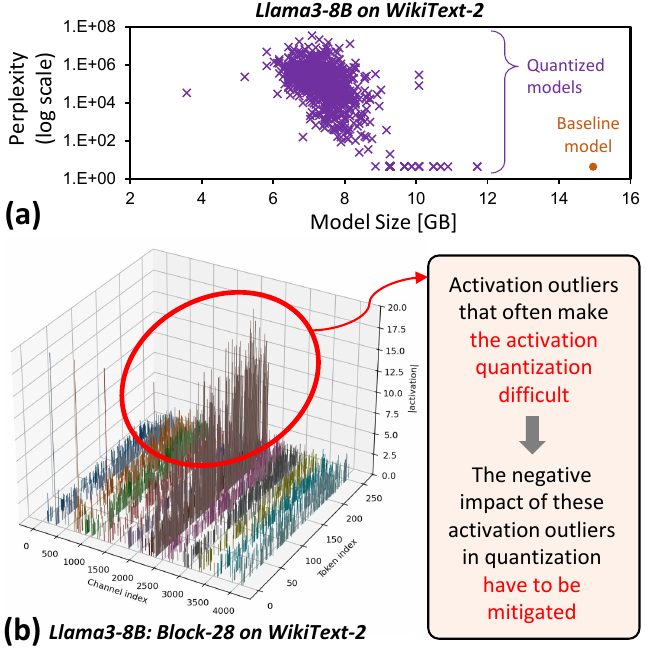}
\vspace{-0.6cm}
\caption{\textbf{(a)} Our experimental results in exploring the performance (perplexity) and memory footprints of quantized Llama3-8B models considering different combinations of weight precision levels across the attention blocks on the WikiText-2 dataset~\cite{Ref_Merity_WikiText_ICLR17}.
\textbf{(b)} Illustration of activation outliers from Llama3-8B at Block-28 on WikiText-2 under a baseline setting (no quantization).}
\label{Fig_CaseStudy}
\vspace{-0.3cm}
\end{figure}

\vspace{-0.2cm}
\subsection{A Case Study and Associated Research Challenges}
\label{Sec_Intro_Challenges}

To show the potentials and challenges in quantizing LLMs, we conduct an experimental case study.
Here, we perform a design space exploration (DSE) to investigate the performance (i.e., perplexity) and memory footprints of quantized Llama3-8B models considering different combinations of weight precision levels across attention blocks.
This case study is conducted on the WikiText-2 dataset~\cite{Ref_Merity_WikiText_ICLR17}, whose detailed experimental setup will be discussed in Section~\ref{Sec_EvalMethod}.
Experimental results are presented in Fig.~\ref{Fig_CaseStudy}, from which we observe that different combinations of weight precision levels (i.e., mixed-precision levels) across the attention blocks potentially lead to pareto-optimal quantized models that preserve high performance with significant memory savings from the baseline model (FP16). 
From this observation, we also draw the associated research challenges below.  
\begin{itemize}[leftmargin=*]
    \item The quantization method should find the appropriate weight precision levels across the attention blocks to ensure the preservation of high performance. 
    Otherwise, the quantization will lead to significant performance degradation.
    \item The quantization method should effectively handle different network architectures due to the non-conventional attention structures that  exist in emerging LLMs. 
    \item The negative impact of activation outliers in quantization have to be mitigated. 
\end{itemize}

\subsection{Our Novel Contributions}
\label{Sec_Intro_Novelty}

\begin{figure}[h]
\vspace{-0.2cm}
\centering
\includegraphics[width=\linewidth]{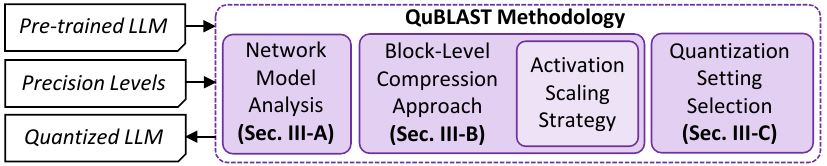}
\vspace{-0.6cm}
\caption{Our novel contributions in this paper.}
\label{Fig_Novelty}
\end{figure}

To address the targeted research problem and its associated challenges, we propose \textit{\textbf{QuBLAST}, a novel \underline{Qu}antization methodology that performs \underline{B}lock-\underline{L}evel compression approach with \underline{A}ctivation \underline{S}caling s\underline{T}rategy for pre-trained LLMs}.
To achieve this, our QuBLAST employs the following key steps (see the overview in Fig.~\ref{Fig_Novelty} and the methodology in Fig.~\ref{fig:qublm_overview}). 
\begin{itemize}[leftmargin=*]
    \item \textbf{Network Model Analysis (Section~\ref{Sec_QuBLM_NetAnalysis}):} 
    It aims to investigate the structure of the pre-trained LLM and determine the model hierarchy, including the architecture of its attention blocks. 
    This step is important for identifying the appropriate quantization approach to take in the next step.
    \item \textbf{Block-Level Compression (Section~\ref{Sec_QuBLM_BlockQuant}):}
    It investigates the sensitivity profile for each attention block under weight quantization using the cross-entropy loss analysis.  
    Then, it performs precision lockout for each attention block based on the loss changes and the user-defined acceptable loss threshold for pooling candidates of quantization settings. 
    If activation compression is required, then activation scaling is employed to control the range of activation values.
    \item \textbf{Quantization Setting Selection (Section~\ref{Sec_QuBLM_QuantSelect}):}
    It selects the final quantization setting/configuration for the model by considering the performance (e.g., perplexity), memory footprint, and user-defined acceptable performance threshold. 
\end{itemize}

\textbf{Key Results:}
We implement the QuBLAST methodology on PyTorch, and run it on the Nvidia A100 GPU machine with 40GB memory. 
Experimental results show that, QuBLAST effectively reduces the memory footprint of investigated models (i.e., Qwen3-8B~\cite{qwen3}, Llama3-8B~\cite{llama3}, Mistral v0.1-8B~\cite{mistral7b}, and Falcon H1R-7B~\cite{falconllm}) by 40\%-45.2\%, while maintaining high performance within 5\% increased perplexity from the original non-quantized models on WikiText-2 and WikiText-103~\cite{Ref_Merity_WikiText_ICLR17}. 
When the activation is quantized, our QuBLAST achieves 42.4\%-48.20\% memory savings while consistently preserving the performance within 2\% increased perplexity from the original non-quantized models on WikiText-2.

\section{Preliminaries}
\label{Sec_Prelim}

In recent years, LLM architectures have evolved to improve the performance, reduce computational complexity, decrease the memory usagae, and increase task flexibility. 
Therefore, in this work, we consider four state-of-the-art LLMs with different attention architectures, including Qwen3-8B~\cite{qwen3}, Llama3-8B~\cite{llama3}, Mistral v0.1-7B~\cite{mistral7b}, as well as Falcon H1R-7B~\cite{falconllm}. 
We specifically select these parameter sizes, as they enable a faster search space for suitable quantization schemes and are representative of models deployed on embedded platforms, which benefit greatly from quantization.

\vspace{-0.2cm}
\subsection{Quantization Schemes}

For quantizing any given LLMs, there are two main options, i.e., \textit{Quantization-aware Training (QAT)} and \textit{Post-training Quantization}. 
QAT involves quantizing the model during the training phase, thereby incurring high computational time, memory cost, and power/energy consumption associated with training. 
Meanwhile, PTQ is employed post-training, hence incurring less overheads as compared to QAT.
Moreover, many studies have shown that PTQ can preserve high performance close to the original non-quantized model. 
Therefore, in this work, we choose to utilize PTQ scheme.
Furthermore, we employ a simulated quantization approach to obtain representative results of the performance and memory statistics expected from deployment while enabling faster DSE. 

\begin{figure*}[t]
    \centering
    \includegraphics[width=\linewidth]{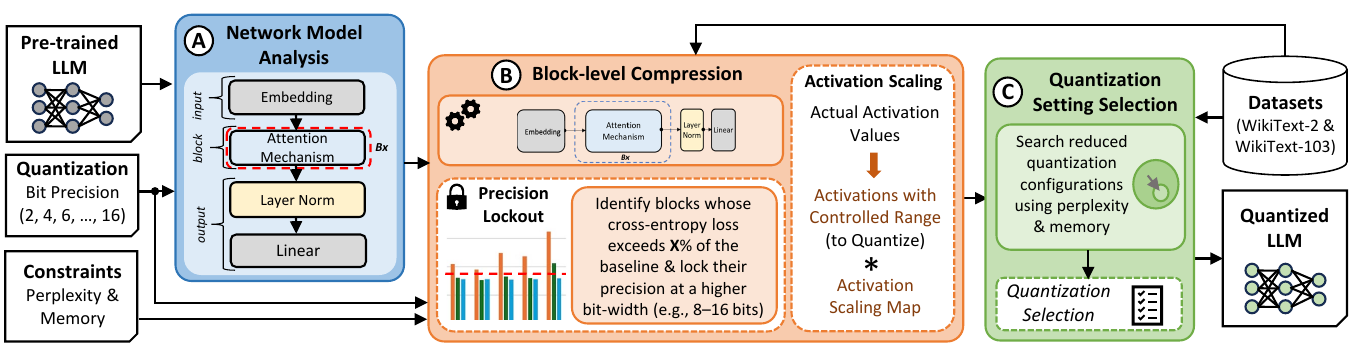}
    \vspace{-0.7cm}
    \caption{Our proposed QuBLAST methodology, showing its key steps: (A) network model analysis, (B) block-level compression, including sensitivity analysis, precision lockout, and activation scaling, as well as (C) quantization setting selection.}
    \label{fig:qublm_overview}
\end{figure*}

\vspace{-0.2cm}
\subsection{LLM Architecture Variants}
\label{Sec_Prelim_ArchVar}

Recent LLM advancements have introduced diverse network architectures, such as the following.
\begin{itemize}[leftmargin=*]
    \item Falcon H1R-7B~\cite{falconllm} utilizes the hybrid Mamba (i.e., state space model) and the self-attention configuration, as shown in \circledB{A} in Fig.~\ref{Fig_BlockGeneralization}.
    \item Mistral v0.1-7B~\cite{mistral7b} performs rotation operations on Query \& Key (QK) values before they reach the attention head.
    \item Qwen3~\cite{qwen3} utilizes a FlashAttention structure to minimize computations. In addition, several versions of Qwen3 are available, e.g., a simplified ``dense'' edition and diversified ``Mixture-of-Experts'' editions; see \circledB{B} in Fig.~\ref{Fig_BlockGeneralization}
\end{itemize} 

These different attention architecture variants further impose additional challenges related to how to efficiently quantize the LLMs.
To address this, we propose the QuBLAST methodology, which is discussed in Section~\ref{Sec_QuBLM}.

\section{The QuBLAST Methodology}
\label{Sec_QuBLM}

Our QuBLAST methodology employs three stages; see Fig.~\ref{fig:qublm_overview}.
First, \textit{Network Model Analysis} examines the architecture of the target LLM and identifies repeated attention blocks as quantization units. 
Second, \textit{Block-Level Compression with Precision Lockout} evaluates each block's tolerance under various bit-widths using cross-entropy loss, and excludes configurations that cause excessive degradation.
This stage also considers activation scaling to control the activation values for enabling activation quantization while preserving high performance.  
Finally, \textit{Quantization Setting Selection} searches over the remaining candidates to select a mixed-precision scheme that maximizes memory savings while satisfying the user-defined performance constraint.

\vspace{-0.2cm}
\subsection{Network Model Analysis}
\label{Sec_QuBLM_NetAnalysis}

\textbf{Model Hierarchy:} 
To apply QuBLAST to the given LLM, we first analyze the model architecture to identify the repeated attention blocks that serve as basic quantization units. 
Specifically, we inspect the model hierarchy and categorize layers into three types: \textit{embedding (input)}, \textit{attention}, and \textit{output layers}. 
Since embedding and output layers are more sensitive to quantization, we exclude them from the search and retain them at full precision. 
Therefore, only the attention blocks are considered as candidates for quantization. 

\begin{figure}[t]
    \centering
    \includegraphics[width=\linewidth]{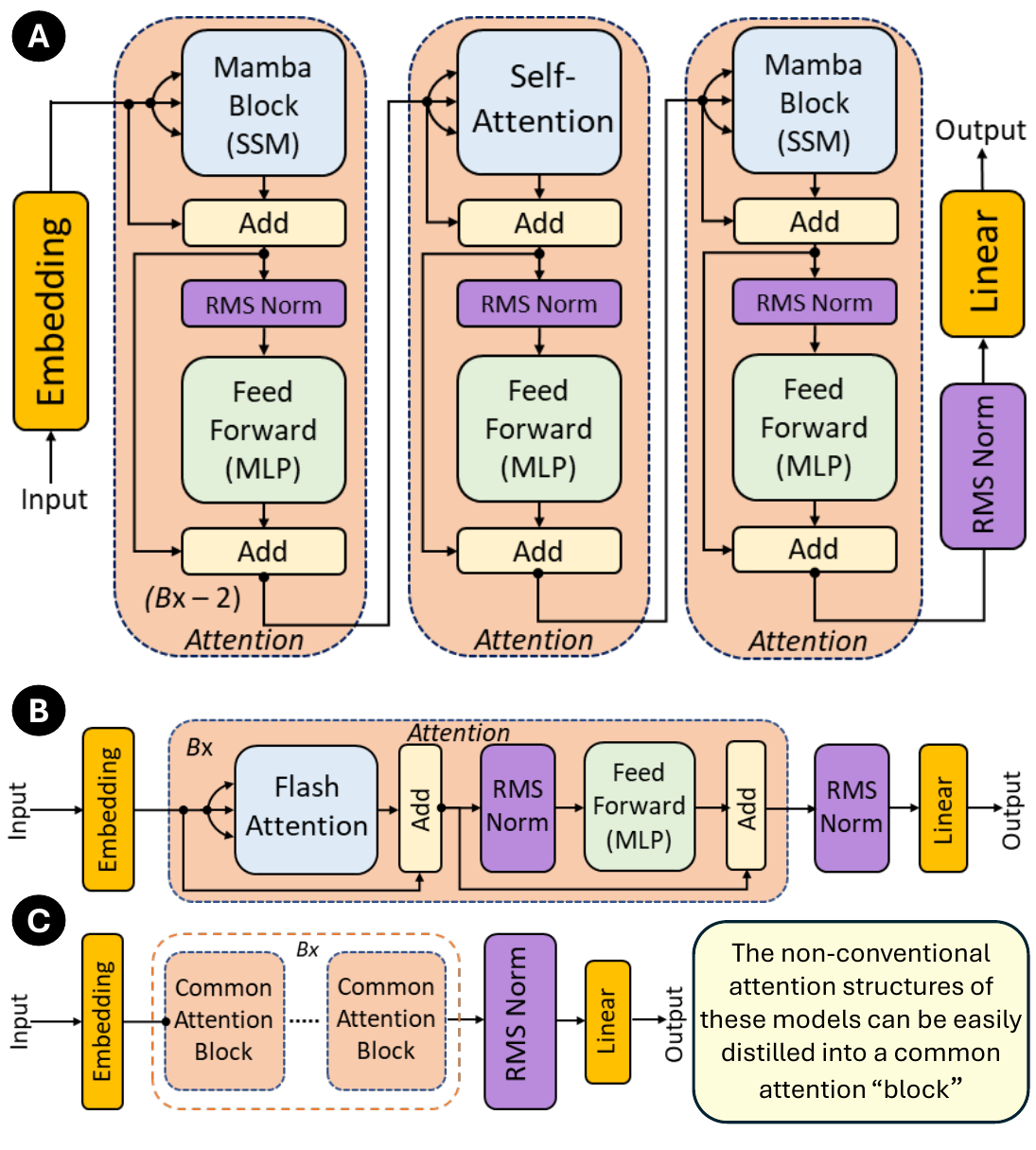}
    \vspace{-0.8cm}
    \caption{(a) The state space model (SSM)-based structure of Falcon H1R-7B. (b) The FlashAttention in Qwen3-8B. (c) Our proposed ``common attention block'' (CAB) denotes how different attentions architectures from different models are distilled into a common block.}
    \label{Fig_BlockGeneralization}
\end{figure}

\smallskip
\textbf{Common Attention Blocks (CABs):} 
As discussed in Section~\ref{Sec_Prelim_ArchVar} and shown by \circledB{A} and \circledB{B} in Fig.~\ref{Fig_BlockGeneralization}, there are different architectural variants for attention blocks, which impose new challenges for performing effective quantization.  
To address this, \textit{we propose to identify the attention blocks as clusters of repeating layers called ``common attention blocks (CABs)'' for the model}, as illustrated in \circledB{C} in Fig.~\ref{Fig_BlockGeneralization}. 
Although these layers appear to be simple recurring copies, each layer plays a different role in the whole LLM pipeline. 
Therefore, if uniform quantization is deployed, significant performance degradation at lower bit-precision levels may occur since the sensitive layers are not accommodated.
This limitation necessitates a thorough analysis of all the layers in a model to determine their sensitivity under quantization. 
While a sensitivity analysis of each and every layer would result in a more comprehensive quantization schema, these LLMs contain hundreds of layers. 
For example, testing the Qwen3-8B with over 250 linear layers against a variety of bit ranges (e.g., {2, 4, 6, 8, 10}) would result in a significant number of combinations to be tested. 
\textit{To address these expensive analysis overheads, we propose \textbf{a  block-level quantization approach}, where we perform the analysis at higher level with the aid of our proposed common attention blocks (CABs)}.

\subsection{Block-Level Compression}
\label{Sec_QuBLM_BlockQuant}

\subsubsection{\textbf{Weight Quantization}}

This stage first aims to profile the sensitivity of each block under quantization, and leverage this information to constrain the pool of quantization configuration candidates for the subsequent configuration search process.
Evaluating all possible configurations directly is computationally prohibitive, so we first identify which bit-widths cause unacceptable degradation for each block and exclude them from consideration. 
This is achieved through two steps, i.e., \textit{\textbf{cross-entropy loss analysis}} and \textit{\textbf{precision lockout}}, as detailed in Algorithm~\ref{alg:block_sensitivity}.
Note, we adopt the cross-entropy loss analysis rather than perplexity as the sensitivity metric. 
The reason is that, when degradation is significant, perplexity tends to over-inflate, obscuring relative comparisons across blocks. 
The cross-entropy loss is computed using Eq.~\ref{Eq_CrossEntropy}, where $N_T$ denotes the token number in the sequence and $P(w_i \mid w_{<i})$ denotes the predicted probability of token $w_i$ given preceding context.

\begin{equation}
    \mathcal{L} = -\frac{1}{N_T} \sum_{i=1}^{N_T} \log P(w_i \mid w_{<i})
    \label{Eq_CrossEntropy}
\end{equation}

\begin{algorithm}[t]
\footnotesize
\caption{Block-level Weight Quantization}
\label{alg:block_sensitivity}
\begin{algorithmic}[1]
\renewcommand{\algorithmicrequire}{\textbf{INPUT:}}
\renewcommand{\algorithmicensure}{\textbf{OUTPUT:}}

\REQUIRE (1) Pre-trained LLM model (\texttt{Net}) with blocks $\mathcal{B}=\{b_1,\ldots,b_M\}$;\\
(2) Calibration dataloader \texttt{D};\\
(3) Candidate bit-width set $\mathcal{Q}=\{q_1,\ldots,q_K\}$;\\
(4) Maximum number of evaluation batches $N_{\text{eval}}$;\\
\ENSURE (1) Block-wise loss table $\mathcal{L}[b][q]$;\\
(2) Block-wise loss increase table $\Delta\mathcal{L}[b][q]$;\\

\smallskip
\textbf{BEGIN} \\
\smallskip
\textbf{Process}: \\
\STATE Evaluate baseline loss: $\mathcal{L}_{\text{base}} \leftarrow \texttt{EvaluateLoss}(\texttt{Net}, \texttt{D}, N_{\text{eval}})$
\STATE Initialize loss tables $\mathcal{L}$ and $\Delta\mathcal{L}$ as empty

\FOR{each block $b \in \mathcal{B}$}
    \FOR{each bit-width $q \in \mathcal{Q}$}
        \IF{$q$ corresponds to full precision}
            \STATE $\mathcal{L}[b][q] \leftarrow \mathcal{L}_{\text{base}}$
        \ELSE
            \STATE Temporarily quantize block $b$ to bit-width $q$
            \STATE $\ell \leftarrow \texttt{EvaluateLoss}(\texttt{Net}, \texttt{D}, N_{\text{eval}})$
            \STATE Restore block $b$ to full precision
            \STATE $\mathcal{L}[b][q] \leftarrow \ell$
        \ENDIF
        \STATE $\Delta\mathcal{L}[b][q] \leftarrow \mathcal{L}[b][q] - \mathcal{L}_{\text{base}}$
    \ENDFOR
\ENDFOR

\STATE \textbf{return} $\mathcal{L}$ and $\Delta\mathcal{L}$;\\
\textbf{END}
\end{algorithmic}
\end{algorithm}
\setlength{\textfloatsep}{6pt}

\textbf{Cross-Entropy Loss Analysis:} 
As shown in Algorithm~\ref{alg:block_sensitivity}, we quantize each block independently using the candidate bit-widths $q \in \{2, 4, 6, 8, 10, 16\}$, while keeping all other blocks at full precision. 
For each block-bitwidth pair, we compute the cross-entropy loss $\mathcal{L}[b][q]$ and the increase $\Delta\mathcal{L}[b][q]$ relative to the baseline. This produces a sensitivity profile, which shows how each block responds to quantization.

\smallskip
\textbf{Precision Lockout:} 
We exclude bit-widths that cause excessive degradation per block. 
If $\Delta\mathcal{L}[b][q]$ exceeds a user-defined acceptable loss threshold $T_{\text{loss}}$ (e.g., 5\% above baseline), bit-width $q$ is locked-out for block $b$. 
For instance, if block-0 exceeds the threshold at 2-bit and 4-bit, the minimum bit-width for that block-0 becomes 6 bits. 
This filtering significantly reduces the pool size of configuration candidates while protecting sensitive blocks from aggressive quantization.

\smallskip
\subsubsection{\textbf{Activation Quantization}}
\begin{figure}[t]
    \centering
    \includegraphics[width=\linewidth]{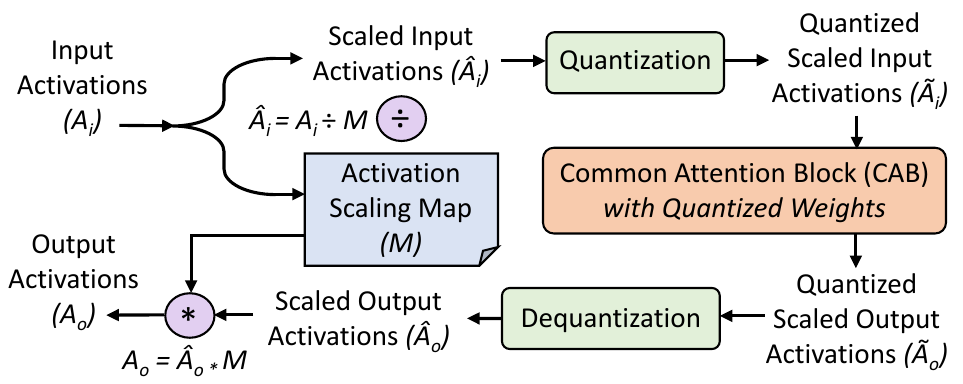}
    \vspace{-0.6cm}
    \caption{Our proposed activation scaling strategy.}
    \label{Fig_AttentionScaling}
\end{figure}

As observed by many previous works~\cite{Ref_Xiao_SmoothQuant_ICML23, Ref_Qiu_QLLM_ICLR24, Ref_Shao_OmniQuant_ICLR24, Ref_Liu_SpinQuant_ICLR25}, quantizing activations is very challenging as activation outliers can degrade the representation capability of quantization for small yet majority values.
Therefore, we also develop an efficient technique to mitigate the negative impact of activation outliers on the performance, called \textit{\textbf{an activation scaling strategy}}.
Its key idea is \textit{to scale-down the actual input activation values ($A_i$) in each attention block with a crafted scaling map ($M$), thereby the scaled activation values ($\hat{A}_i$) are controlled within a certain range that can be quantized with good representation capability}; see the illustration in Fig.~\ref{Fig_AttentionScaling}.

\subsection{Quantization Setting Selection}
\label{Sec_QuBLM_QuantSelect}

After precision lockout, we evaluate the remaining quantization configuration candidates to identify the best memory-performance trade-off. 
For each configuration, we quantize the model accordingly, compute its performance (e.g., perplexity) on the calibration dataset, and estimate its memory footprint. 
Configurations that meet the user-defined acceptable performance threshold ($T_{\text{perf}}\%$) are marked as viable. 
Algorithm~\ref{alg:quant_search} describes this process in details.
Meanwhile, memory footprint is estimated by separating parameters into two groups, i.e., \textit{non-quantized parameters} (i.e., embedding and output layers) stored at 16-bit precision, and \textit{quantizable block parameters} stored at their assigned bit-widths. 
The total memory footprint is the sum of these components. 
For performance in terms of perplexity (which is the exponential of cross-entropy loss) can be calculated using Eq.~\ref{Eq_Perplexity}.
\begin{equation}
    \text{PPL} = \exp(\mathcal{L}) = \exp\left(-\frac{1}{N_T} \sum_{i=1}^{N_T} \log P(w_i \mid w_{<i})\right)
    \label{Eq_Perplexity}
\end{equation}
From the viable configurations, we then select the one with the lowest memory footprint as the final quantization scheme. 
In cases where multiple viable configurations found, the user may select the ideal scheme.

\begin{algorithm}[t]
\footnotesize
\caption{Quantization Setting Selection}
\label{alg:quant_search}
\begin{algorithmic}[1]
\renewcommand{\algorithmicrequire}{\textbf{INPUT:}}
\renewcommand{\algorithmicensure}{\textbf{OUTPUT:}}
\REQUIRE (1) Pre-trained LLM Model (\texttt{Net}) with blocks $B = \{b_1, \ldots, b_n\}$; \\
\hspace*{2.3em}(2) Calibration dataloader $\mathcal{D}$; \\
\hspace*{2.3em}(3) Quantization Configuration list $Q$; \\
\hspace*{2.3em}(4) Max Perplexity Degradation Threshold $T_{\text{perf}}\%$; \\
\ENSURE (1) List of valid candidates \texttt{qNet}; \\
\hspace*{2.3em}(2) List of statistics \texttt{qStats}; \\
\textbf{BEGIN}
\STATE $PPL_{base} \leftarrow$ \texttt{EvaluatePerplexity}(\texttt{Net}, $Q[0]$) // assume $Q[0]$ is FP16
\STATE Initialize \texttt{qNet}, \texttt{qStats} $\leftarrow$ empty
\STATE Initialize $net\_index \leftarrow 0$
\FOR{each config $c \in Q$}
    \STATE $\texttt{qNet}_{tmp} \leftarrow \texttt{Copy}(\texttt{Net})$
    \FOR{each block $b \in B$}
        \STATE $\texttt{qNet}_{tmp}[b] \leftarrow \texttt{Quantize}(b, Q[c][b])$
    \ENDFOR
    \STATE \texttt{qStat}$_{tmp}$.mem $\leftarrow$ \texttt{CalculateMemory}(\texttt{Net}, $\texttt{qNet}_{tmp}, c$)
    \STATE \texttt{qStat}$_{tmp}$.ppl $\leftarrow$ \texttt{EvaluatePerplexity}($\texttt{qNet}_{tmp}, \mathcal{D}$)
    \IF{\texttt{qStat}$_{tmp}$.ppl $\le PPL_{base} \times (1 + \frac{T_{\text{perf}}}{100})$}
        \STATE $\texttt{qNet}[net\_index] \leftarrow \texttt{qNet}_{tmp}$
        \STATE $\texttt{qStats}[net\_index] \leftarrow \texttt{qStat}_{tmp}$
        \STATE $net\_index \leftarrow net\_index + 1$
    \ENDIF
\ENDFOR
\STATE \textbf{return} \texttt{qNet}, \texttt{qStats}; \\
\textbf{END}
\end{algorithmic}
\end{algorithm}
\setlength{\textfloatsep}{6pt}

\color{black}

\section{Evaluation Methodology}
\label{Sec_EvalMethod}

\begin{figure}[h]
    \centering
    \includegraphics[width=\linewidth]{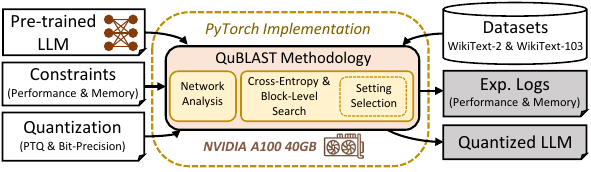}
    \caption{The experimental setup in this work.}
    \label{Fig_ExpSetup}
\end{figure}

To evaluate our QuBLAST methodology, we first implement it in PyTorch and then conduct experiments on a single NVIDIA A100 GPU with 40 GB memory, as illustrated in Fig.~\ref{Fig_ExpSetup}. 
Both sensitivity analysis and configuration search are performed on this device without model fine-tuning or retraining, as QuBLAST is a purely post-training method.

\begin{figure*}
    \centering
    \includegraphics[width=0.98\linewidth]{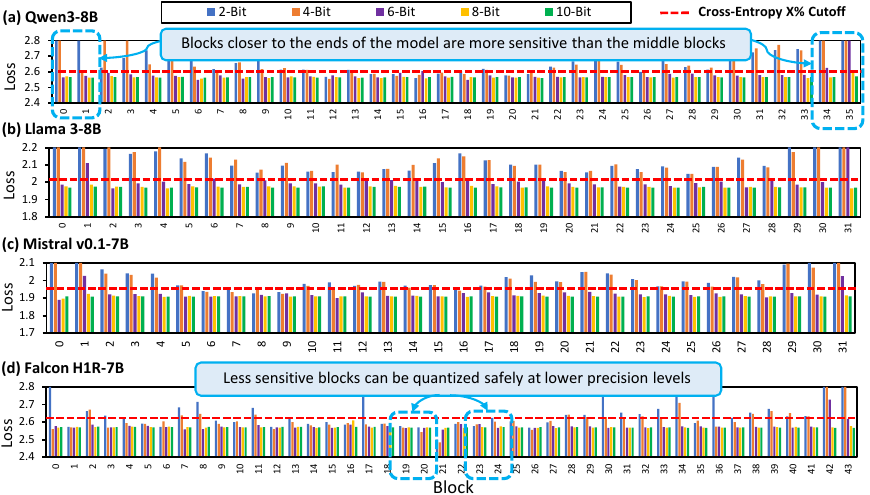}
    \vspace{-0.3cm}
    \caption{Cross-entropy loss across bit-precisions \{2, 4, 6, 8, 10\} for (a) Qwen3-8B, (b) Llama 3-8B, (c) Mistral v0.1-7B and (d) Falcon H1R-7B.}
    \label{fig:BlockSensitivity}
    \vspace{-0.3cm}
\end{figure*}

\smallskip
\textbf{Models and Datasets:}
We evaluate QuBLAST on four prominent pre-trained LLMs: Qwen3-8B~\cite{qwen3}, Llama3-8B~\cite{llama3}, Mistral v0.1-7B~\cite{mistral7b}, and Falcon H1R-7B~\cite{falconllm}, using their original publicly available checkpoints. 
We use the WikiText-2 dataset for the block sensitivity analysis in the first stage, and both the WikiText-2 and WikiText-103 datasets~\cite{Ref_Merity_WikiText_ICLR17} for the quantization configuration search in the second stage to assess generalization across datasets.

\smallskip
\textbf{Block Sensitivity Analysis:}
We aim to quantize each block independently considering the following bit-width candidates: $\{2, 4, 6, 8, 10, 12, 14, 16\}$, while keeping all other blocks at full precision, where 16-bit serves as the baseline. We measure sensitivity as the relative increase in the cross-entropy loss $\Delta\mathcal{L}$ compared to the baseline. Here, two user-defined acceptable loss thresholds $T_{\text{loss}}$ are evaluated: 2\% and 5\% above baseline. 
Block and bit-width combinations exceeding the threshold are excluded from the second-stage search.

\smallskip
\textbf{Mixed-Precision Configuration Search and Metrics:}
We generate a set of configuration candidates for each model, including the uniform-precision and randomly-sampled mixed-precision assignments. 
After applying the precision lockout constraints from the first stage, we evaluate the remaining candidates using perplexity on both the WikiText-2 and WikiText-103 datasets. 
A configuration is considered valid if its perplexity increase stays within 5\% of the baseline. 
Then, we estimate memory footprint analytically based on parameter counts and assigned bit-widths, as described in Section~\ref{Sec_QuBLM_QuantSelect}.

\section{Results and Discussion}
\label{Sec_Results}

\begin{figure*}
    \centering
    \includegraphics[width=1\linewidth]{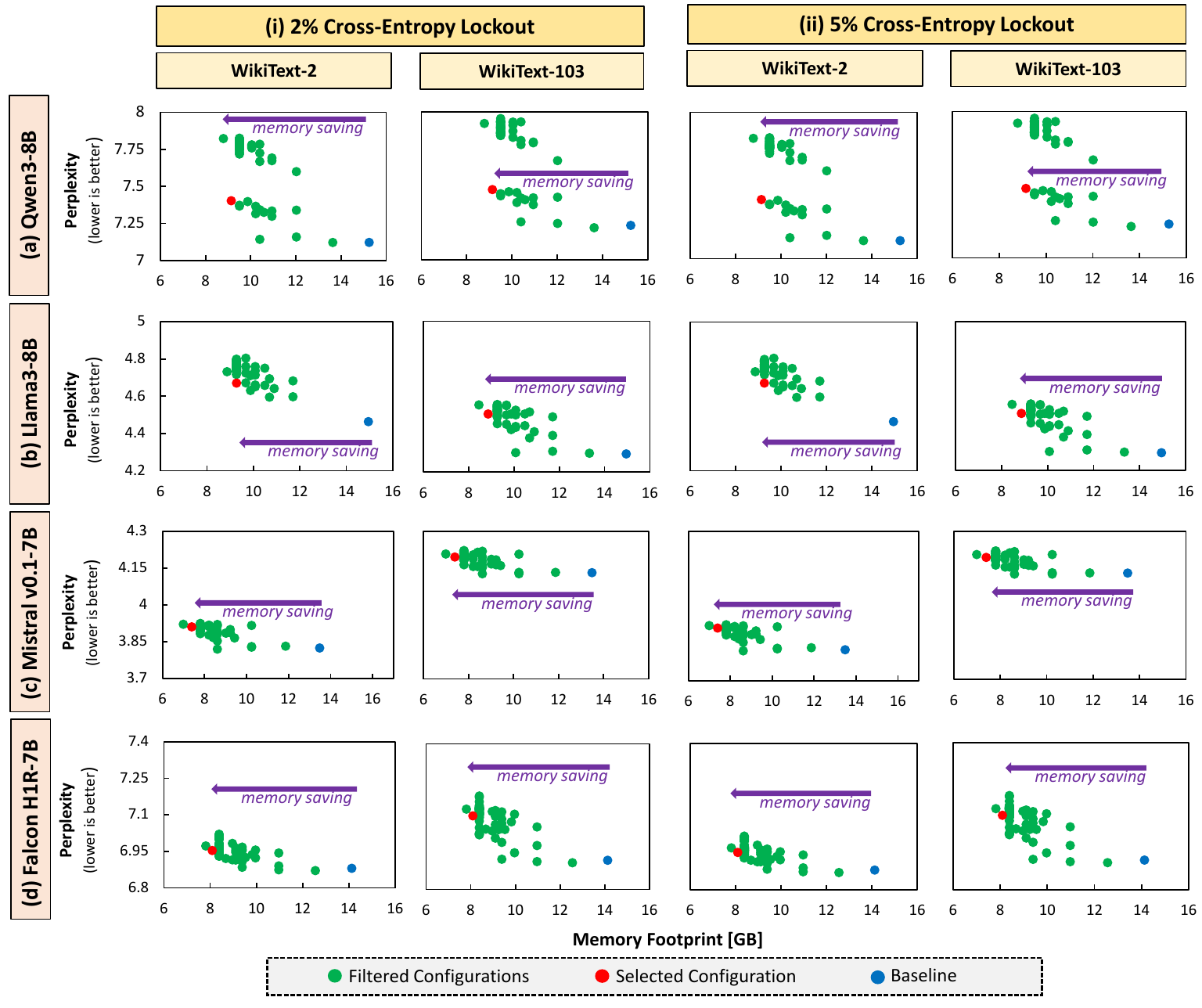}
    \vspace{-0.7cm}
    \caption{Evaluation of the quantization setting candidates in terms of perplexity vs. model size for \textbf{(a)} Qwen3-8B, \textbf{(b)} Llama 3-8B, \textbf{(c)} Mistral v0.1-7B and \textbf{(d)} Falcon H1R-7B. 
    The displayed configuration candidates are selected after precision lockout at \textbf{(i)} 2\% and \textbf{(ii)} 5\% acceptable loss thresholds.}
    \label{fig:QuantSetting}
\end{figure*}

\subsection{Comparison Analysis}

Table~\ref{tab:sota_comparison} compares different post-training quantization baselines and reveals a clear trade-off between compression aggressiveness and performance preservation. Conservative schemes, such as uniform 8-bit quantization with activation scaling, and prior 8-bit PTQ baselines where available, retain model quality well because they operate in a relatively safe quantization regime. However, this safety also limits their ability to fully exploit the available compression headroom. At the other end of the spectrum, more aggressive low-bit baselines, including uniform 4-bit quantization and GPTQ-style settings, highlight the fragility of applying a single precision policy across all blocks. Their behavior becomes increasingly unstable when quantization is pushed uniformly to lower precision levels. In this context, the main advantage of QuBLAST is not merely that it targets a lower nominal precision. Instead, it avoids the one-size-fits-all assumption by assigning higher precision only to quantization-sensitive blocks while compressing more tolerant blocks more aggressively. As a result, QuBLAST reaches a more favorable memory, performance trade-off under a bounded performance degradation budget, suggesting that the main limitation of low-bit PTQ lies in uniformity rather than in low precision itself.

\begin{table}[t]
\centering
\caption{Comparison on different PTQ methods. $\Delta$PPL denotes relative perplexity increase (\%) from each method's own FP16 baseline on WikiText-2. Results with $^\dagger$/$^\ddagger$ are from prior papers.}
\label{tab:sota_comparison}
\renewcommand{\arraystretch}{1.15}
\resizebox{\columnwidth}{!}{%
\begin{tabular}{llcccc}
\toprule
& & \multicolumn{2}{c}{\textbf{Llama3-8B}} & \multicolumn{2}{c}{\textbf{Mistral v0.1-7B}} \\
\cmidrule(lr){3-4} \cmidrule(lr){5-6}
\textbf{Method} & \textbf{Prec.} & $\Delta$PPL$\downarrow$ & Mem.$\downarrow$ & $\Delta$PPL$\downarrow$ & Mem.$\downarrow$ \\
\midrule
Uniform W8 + AS & W8 & 1.85\% & $\sim$50\% & 0.68\% & $\sim$50\% \\
Uniform W4 + AS & W4 & $\gg$100\% & $\sim$75\% & $\gg$100\% & $\sim$75\% \\
\midrule
SmoothQuant$^\dagger$ & W8A8 & -- & $\sim$50\%$^\S$ & 0.46\% & $\sim$50\%$^\S$ \\
GPTQ$^\ddagger$ & W4 & 18.0\% & $\sim$75\%$^\S$ & 59.3\% & $\sim$75\%$^\S$ \\
SpinQuant-nohad$^\ddagger$ & W4A8 & 9.8\% & $\sim$75\%$^\S$ & 5.6\% & $\sim$75\%$^\S$ \\
\midrule
\textbf{QuBLAST (Ours)} & \textbf{Mixed} & \textbf{4.64\%} & \textbf{40.7\%} & \textbf{2.27\%} & \textbf{45.2\%} \\
\bottomrule
\end{tabular}%
}
\end{table}

Another important observation from Table~\ref{tab:sota_comparison} is that, QuBLAST exhibits stronger cross-model consistency than fixed-rule baselines. Methods built around a single global quantization recipe may appear effective for one architecture but degrade noticeably on another, indicating that their behavior is tightly coupled to model-specific characteristics. In contrast, QuBLAST adapts its mixed-precision configuration to the block-level sensitivity profile of each model, making its compression strategy inherently more robust to architectural variation. This property is particularly important in our setting, where the evaluated LLMs do not share the same internal attention structure. \textit{Therefore, the comparison supports the central conclusion of this work: effective post-training quantization for modern LLMs should be sensitivity-aware and architecture-adaptive, rather than globally uniform}. Overall, Table~\ref{tab:sota_comparison} supports the view that block-level mixed-precision quantization is a key mechanism enabling QuBLAST to deliver stable and transferable compression across the compared models while keeping performance degradation tightly controlled.

\subsection{Block Sensitivity Results}
\label{Sec_Results_BlockSensitivity}

Fig.~\ref{fig:BlockSensitivity} presents the block-level quantization sensitivity analysis, which identifies blocks that incur disproportionate cross-entropy loss increases under low-bit quantization, thus motivating the subsequent precision lockout strategy, as discussed in Section~\ref{Sec_QuBLM_BlockQuant}. 
This sensitivity analysis shows clear non-uniform behavior across attention blocks and models, particularly at low bit-widths (i.e., 2–4 bits). 
This indicates that uniform precision assignment does not align well with the quantization behavior of LLMs. 
At these precision levels, a small subset of blocks exhibits large increases in cross-entropy loss, while the majority of blocks remain comparatively stable once moderate bit-widths are used.

Across all evaluated models, most blocks exhibit a stabilization point between 6-bit and 8-bit precision, beyond which further increases in bit-width result in minimal changes in loss. 
Sensitivity is also correlated with block position. Early blocks, especially block-0 to block-2, are consistently among the most sensitive at very low bit-widths across all evaluated models. 
Final blocks also show elevated sensitivity in several cases, though their impact is generally less pronounced than that of the initial blocks. 
In contrast, middle blocks tend to be more tolerant and typically stabilize earlier, often at 6–8 bits. 
These observations motivate the second stage of our QuBLAST methodology, where block-level sensitivity is used to constrain the mixed-precision search space; this will be discussed further in Section~\ref{Sec_Results_MixedPrecision}.

\subsection{Mixed-Precision Configuration Results}
\label{Sec_Results_MixedPrecision}

Building on the block-level sensitivity analysis from Section~\ref{Sec_Results_BlockSensitivity}, Fig.~\ref{fig:QuantSetting} reports the second-stage mixed-precision search results for all evaluated models, showing the filtered configurations post-precision lockout stage (i.e., testing on 2\% and 5\% of $T_{\text{loss}}$), the final selected configuration, and the full-precision baseline considering both the WikiText-2 and WikiText-103 datasets. 
These results show that, our QuBLAST performs a constrained mixed-precision configuration search.
Here, only non-sensitive blocks are allowed to operate at low bit-widths, while sensitive blocks are locked at higher precision. 
This substantially reduces the search space and enables aggressive quantization without violating the performance constraint (e.g., perplexity score). 
The second-stage results show that, QuBLAST consistently identifies configurations that achieve better memory–performance trade-offs than uniform-precision baselines. 
For Qwen3-8B and Llama3-8B, uniform low-bit quantization fails to remain within the 5\% perplexity threshold, whereas QuBLAST selects mixed-precision configurations that satisfy the constraint while achieving meaningful memory reduction. 
For Mistral v0.1-7B and Falcon H1R-7B, although uniform-precision baselines fall within the threshold, QuBLAST identifies alternative configurations that achieve comparable or lower perplexity with similar or only slightly reduced memory savings. 

As summarized in Table I, the final selected configurations achieve memory reductions in the range of 40–45\% across all evaluated models while keeping perplexity increases under 5\% relative to the full-precision baseline.
Specifically, QuBLAST saves memory footprints by 40\% for Qwen3-8B, 40.7\% for Llama3-8B, 45.2\% for Mistral v0.1-7B, and 42.7\% for Falcon H1R-7B.
Meanwhile, Table~\ref{Tab_PPL} shows the increases in perplexity as a result of memory savings from Table~\ref{Tab_Memory}, with percentages as low as 1.06\% (for Falcon H1R-7B) for a roughly 42\% reduction in memory usage. 
These results demonstrate that QuBLAST consistently attains 40–45\% memory reduction across models while remaining within the prescribed perplexity constraint.
These results also confirm that the sensitivity under quantization is highly non-uniform across blocks, and exploiting this non-uniformity enables effective mixed-precision model selection.

\begin{table}[h]
\caption{Memory Savings (in \%) for different models on WikiText datasets. Results for Acceptable Loss Thresholds of 2\% or 5\% respectively. NOTE: Higher is Better.}
\footnotesize
\centering
\begin{tabular}{|c|c|c|}
\hline
  \textbf{Model} & \textbf{Dataset} & \begin{tabular}[c]{@{}c@{}} \textbf{Loss Thresholds } \\ \textbf{($T_{\text{loss}}$ = 2\% or 5\%)} \end{tabular} \\ \hline \hline
  Qwen3-8B\cite{qwen3} & \begin{tabular}[c]{@{}c@{}} WikiText-2 \\ WikiText-103 \end{tabular} & \begin{tabular}[c]{@{}c@{}} 40.0\%  \end{tabular} \\ \hline
  Llama 3-8B\cite{llama3} & \begin{tabular}[c]{@{}c@{}} WikiText-2 \\ WikiText-103 \end{tabular} & \begin{tabular}[c]{@{}c@{}} 40.7\%  \end{tabular} \\ \hline
  Mistral v0.1-7B\cite{mistral7b} & \begin{tabular}[c]{@{}c@{}} WikiText-2 \\ WikiText-103 \end{tabular} & \begin{tabular}[c]{@{}c@{}} 45.2\%  \end{tabular} \\ \hline
  Falcon H1R-7B\cite{falconllm} & \begin{tabular}[c]{@{}c@{}} WikiText-2 \\ WikiText-103 \end{tabular} & \begin{tabular}[c]{@{}c@{}} 42.7\%  \end{tabular} \\ \hline
\end{tabular}
\label{Tab_Memory}
\end{table}

\begin{table}[h]
\caption{Perplexity Increases (in \%) for different models on WikiText datasets. Results for Acceptable Loss Thresholds of 2\% or 5\% respectively. NOTE: Lower is better.}
\footnotesize
\centering
\begin{tabular}{|c|c|c|}
\hline
  \textbf{Model} & \textbf{Dataset} & \begin{tabular}[c]{@{}c@{}} \textbf{Loss Thresholds } \\ \textbf{($T_{\text{loss}}$ = 2\% or 5\%)} \end{tabular} \\ \hline \hline
  Qwen3-8B\cite{qwen3} & \begin{tabular}[c]{@{}c@{}} WikiText-2 \\ WikiText-103 \end{tabular} & \begin{tabular}[c]{@{}c@{}} 3.95\% \\ 3.36\% \end{tabular} \\ \hline
  Llama 3-8B\cite{llama3} & \begin{tabular}[c]{@{}c@{}} WikiText-2 \\ WikiText-103 \end{tabular} & \begin{tabular}[c]{@{}c@{}} 4.64\% \\ 4.94\% \end{tabular} \\ \hline
  Mistral v0.1-7B\cite{mistral7b} & \begin{tabular}[c]{@{}c@{}} WikiText-2 \\ WikiText-103 \end{tabular} & \begin{tabular}[c]{@{}c@{}} 2.27\% \\ 1.54\% \end{tabular} \\ \hline
  Falcon H1R-7B\cite{falconllm} & \begin{tabular}[c]{@{}c@{}} WikiText-2 \\ WikiText-103 \end{tabular} & \begin{tabular}[c]{@{}c@{}} 1.06\% \\ 2.65\% \end{tabular} \\ \hline
\end{tabular}
\label{Tab_PPL}
\end{table}

\subsection{Memory Analysis}

Table~\ref{tab:config_mem} reports the memory saving of all viable quantization configurations under progressively broader compression scopes. A clear pattern emerges from these results. The gains in memory efficiency increase much faster than the expansion of the quantized scope, while the impact on perplexity remains consistently small. Moving from a conservative center only configuration to the full configuration yields roughly four to five times more memory savings across all four models, yet the associated perplexity increase stays marginal. In Llama3 8B, for instance, the center configuration reduces memory usage by 1.22~GB with only a 0.02 increase in perplexity, whereas the full configuration saves 6.09~GB with a perplexity increase of just 0.07. This favorable trend strongly supports the block sensitivity structure leveraged by QuBLAST. Since most middle blocks are highly tolerant to quantization, each newly quantized block contributes meaningful memory reduction while introducing only minimal additional performance loss.

\begin{table}[t]
\centering
\caption{Memory saving of viable quantization configurations on WikiText-2 with activation scaling enabled. Entries marked with $\dagger$ denote configurations produced by QuBLAST.}
\label{tab:config_mem}
\resizebox{\columnwidth}{!}{%
\begin{tabular}{llcccc}
\hline
\textbf{Model} & \textbf{Configuration} & \textbf{PPL} & \textbf{Size} & \textbf{Savings} & \textbf{Savings} \\
 & & & \textbf{[GB]} & \textbf{[GB]} & \textbf{[\%]} \\
\hline
\multirow{5}{*}{Qwen3-8B}
 & FP16 Baseline                          & 8.60  & 15.256 & 0      & 0\%     \\
 & Mixed W8A8/FP16 center$^{\dagger}$     & 8.60  & 14.178 & 1.078  & 7.1\%   \\
 & Mixed W8A8/FP16 wide$^{\dagger}$       & 8.62  & 12.022 & 3.234  & 21.2\%  \\
 & Mixed W8A8/FP16 full$^{\dagger}$       & 8.73  & 9.866  & 5.391  & 35.3\%  \\
 & Uniform W8A8$^{\dagger}$                           & 8.51  & 8.788  & 6.469  & 42.4\%  \\
\hline
\multirow{5}{*}{Llama3-8B}
 & FP16 Baseline                          & 5.50  & 14.958 & 0      & 0\%     \\
 & Mixed W8A8/FP16 center$^{\dagger}$     & 5.51  & 13.739 & 1.219  & 8.1\%   \\
 & Mixed W8A8/FP16 wide$^{\dagger}$       & 5.54  & 11.301 & 3.656  & 24.4\%  \\
 & Mixed W8A8/FP16 full$^{\dagger}$       & 5.57  & 8.864  & 6.094  & 40.7\%  \\
 & Uniform W8A8$^{\dagger}$                           & 5.60  & 8.458  & 6.500  & 43.5\%  \\
\hline
\multirow{5}{*}{Mistral v0.1-7B}
 & FP16 Baseline                          & 4.69  & 13.489 & 0      & 0\%     \\
 & Mixed W8A8/FP16 center$^{\dagger}$     & 4.70  & 12.270 & 1.219  & 9.0\%   \\
 & Mixed W8A8/FP16 wide$^{\dagger}$       & 4.71  & 9.833  & 3.656  & 27.1\%  \\
 & Mixed W8A8/FP16 full$^{\dagger}$       & 4.72  & 7.395  & 6.094  & 45.2\%  \\
 & Uniform W8A8$^{\dagger}$                           & 4.73  & 6.989  & 6.500  & 48.2\%  \\
\hline
\multirow{5}{*}{Falcon H1R-7B}
 & FP16 Baseline                          & 9.76  & 14.129 & 0      & 0\%     \\
 & Mixed W8A8/FP16 center$^{\dagger}$     & 9.73  & 13.268 & 0.862  & 6.1\%   \\
 & Mixed W8A8/FP16 wide$^{\dagger}$       & 9.73  & 11.544 & 2.585  & 18.3\%  \\
 & Mixed W8A8/FP16 full$^{\dagger}$       & 9.71  & 9.821  & 4.309  & 30.5\%  \\
 & Uniform W8A8$^{\dagger}$                           & 9.60  & 7.810  & 6.319  & 44.7\%  \\
\hline
\end{tabular}%
}
\end{table}

Another important finding is that the QuBLAST mixed precision configurations achieve memory savings that are within 3\% to 5\%-age points of uniform W8A8, while consistently delivering lower perplexity. This result shows that sensitivity guided precision allocation is more effective than applying a single bit width uniformly across all blocks. Falcon H1R 7B, despite having 44 blocks and a hybrid Mamba architecture, exhibits the same favorable scaling behavior as the standard transformer models. This suggests that QuBLAST extends beyond conventional transformer designs and remains effective for non standard attention architectures. Taken together, these results show that QuBLAST can identify a precise boundary between compressible and sensitive blocks, allowing the model to approach the maximum achievable memory savings without triggering substantial performance degradation.

\subsection{Impact of the Cross-Entropy Lockout Threshold}

To evaluate the impact of the cross-entropy lockout threshold, QuBLAST was evaluated using 2\% and 5\% loss thresholds during the first-stage block sensitivity analysis. 
Results in Fig.~\ref{fig:QuantSetting} show that, the 2\% loss threshold excludes a larger set of low-bit configurations by locking more blocks at higher precision, resulting in a more constrained search space in the second stage. 
In contrast, the 5\% threshold allows a wider range of candidate configurations to be explored.

Despite these differences in search space size, our QuBLAST converges to the same selected mixed-precision configuration under both lockout thresholds across all evaluated models. 
As shown in Fig.~\ref{fig:QuantSetting} and summarized in Table~\ref{Tab_Memory} and Table~\ref{Tab_PPL}, the selected configurations achieve identical memory savings and remain within the prescribed perplexity constraint regardless of whether a 2\% or 5\% threshold is used. 
This indicates that the final configuration selected by QuBLAST is stable with respect to variations in the sensitivity thresholds. 
Moreover, the two-stage search in QuBLAST is driven by consistent block-level sensitivity patterns.

\subsection{Ablation Study}

Table~\ref{tab:ablation_as} isolates the effect of activation scaling by comparing each quantization configuration with and without per-channel scaling, while keeping all other components of QuBLAST unchanged. The results make one conclusion unmistakably clear. Activation scaling is not simply helpful for post-training quantization, but fundamentally necessary. Without activation scaling, uniform W8A8 quantization causes perplexity to increase catastrophically in the conventional transformer models. In Llama3 8B, perplexity rises from 5.60 to more than 608{,}000. In Mistral v0.1 7B, it increases from 4.73 to over 24{,}000. In Qwen3 8B, it surges from 8.51 to more than 1.3 million. By contrast, once activation scaling is applied, perplexity remains within 2\% of the FP16 baseline under the same precision setting across all four models.

\begin{table}[h]
\centering
\caption{Ablation on activation scaling on WikiText-2. PPL denotes perplexity; lower is better. Entries marked with $\dagger$ denote configurations produced by QuBLAST.}
\label{tab:ablation_as}
\resizebox{\columnwidth}{!}{%
\begin{tabular}{llcc}
\hline
\textbf{Model} & \textbf{Precision} & \textbf{PPL w/ AS} & \textbf{PPL w/o AS} \\
\hline
\multirow{4}{*}{Qwen3-8B}
 & FP16 Baseline                    & 8.60      & 8.60 \\
 & Uniform W8A8$^{\dagger}$                     & 8.51      & 1{,}372{,}634 \\
 & Mixed W8A8/FP16$^{\dagger}$      & 8.72      & 154{,}182 \\
 & Uniform W4A4$^{\dagger}$                     & 22{,}583  & 11{,}901{,}094 \\
\hline
\multirow{4}{*}{Llama3-8B}
 & FP16 Baseline                    & 5.50      & 5.50 \\
 & Uniform W8A8$^{\dagger}$                     & 5.60      & 608{,}566 \\
 & Mixed W8A8/FP16$^{\dagger}$      & 5.56      & 43{,}264 \\
 & Uniform W4A4$^{\dagger}$                     & 964.06    & 601{,}641 \\
\hline
\multirow{4}{*}{Mistral v0.1-7B}
 & FP16 Baseline                    & 4.69      & 4.69 \\
 & Uniform W8A8$^{\dagger}$                     & 4.73      & 24{,}423 \\
 & Mixed W8A8/FP16$^{\dagger}$      & 4.72      & 24{,}421 \\
 & Uniform W4A4$^{\dagger}$                     & 6{,}146   & 21{,}098 \\
\hline
\multirow{4}{*}{Falcon H1R-7B}
 & FP16 Baseline                    & 9.76      & 9.76 \\
 & Uniform W8A8$^{\dagger}$                     & 9.60      & 723 \\
 & Mixed W8A8/FP16$^{\dagger}$      & 9.67      & 173 \\
 & Uniform W4A4$^{\dagger}$                     & 124{,}219 & 4{,}748 \\
\hline
\end{tabular}%
}
\end{table}

This sharp contrast points to a central insight. The main limitation of weight quantization in large language models is not the lower bit width itself, but the way activation outliers amplify rounding errors. Per-channel activation scaling mitigates this effect by shifting part of the quantization difficulty away from activations and onto weights, thereby turning an otherwise unstable quantization setting into a reliable one. Falcon H1R 7B shows a less dramatic degradation without activation scaling at W8A8, likely because its hybrid Mamba architecture produces smoother activation distributions than standard self-attention models. Even so, its perplexity still rises from 9.60 to 723 in the absence of scaling, which confirms that activation scaling remains critical across all evaluated architectures.

\normalcolor
\section{Conclusion}
\label{Sec_Conclude}

In this paper, we propose QuBLAST, a novel block-level and mixed-precision post-training quantization methodology for compressing LLMs. 
QuBLAST uses a two-stage approach. 
First, it performs block-wise sensitivity analysis using cross-entropy loss to identify and lockout low-bit settings for sensitive blocks. 
Then, it searches the reduced configuration space to select a mixed-precision scheme that maximizes memory savings while meeting a fixed perplexity constraint. 
Across four model architectures, Qwen3-8B, Llama3-8B, Mistral v0.1-7B, and Falcon H1R-7B, QuBLAST consistently achieves 40\% to 45.2\% memory reduction while keeping perplexity within 5\% of the full-precision baseline on both the WikiText-2 and WikiText-103 datasets. 
When the activation is quantized, our QuBLAST achieves 42.4\%-48.20\% memory savings while consistently preserving the performance within 2\% increased perplexity from the original non-quantized models on WikiText-2. 
These results show that exploiting block-level sensitivity patterns provides a practical way to compress LLMs more effectively than uniform precision baselines, improving their suitability for memory-constrained deployment.

\section*{Acknowledgment}
This work was partially supported by the NYUAD Center for Center for Cyber Security (CCS), funded by Tamkeen under the NYUAD Research Institute Award G1104. 
Experiments are performed with NYUAD Jubail High Performance Computing. 


\bibliographystyle{IEEEtran}
\bibliography{bibliography}

\end{document}